\documentclass[acmtog,nonacm]{acmart}

\usepackage{booktabs} 

\usepackage{cleveref}

\crefname{figure}{Fig.}{Figs.}
\Crefname{figure}{Fig.}{Figs.}

\crefname{table}{Tab.}{Tabs.}
\Crefname{table}{Tab.}{Tabs.}

\crefname{section}{Sec.}{Secs.}
\Crefname{section}{Sec.}{Secs.}

\crefname{equation}{Eq.}{Eqs.}
\Crefname{equation}{Eq.}{Eqs.}

\usepackage[ruled]{algorithm2e} 

\SetAlFnt{\small}
\SetAlCapFnt{\small}
\SetAlCapNameFnt{\small}
\SetAlCapHSkip{0pt}

\acmJournal{TOG}

\begin{document}

\begin{teaserfigure}
\centering
  \includegraphics[width=1.0\linewidth]{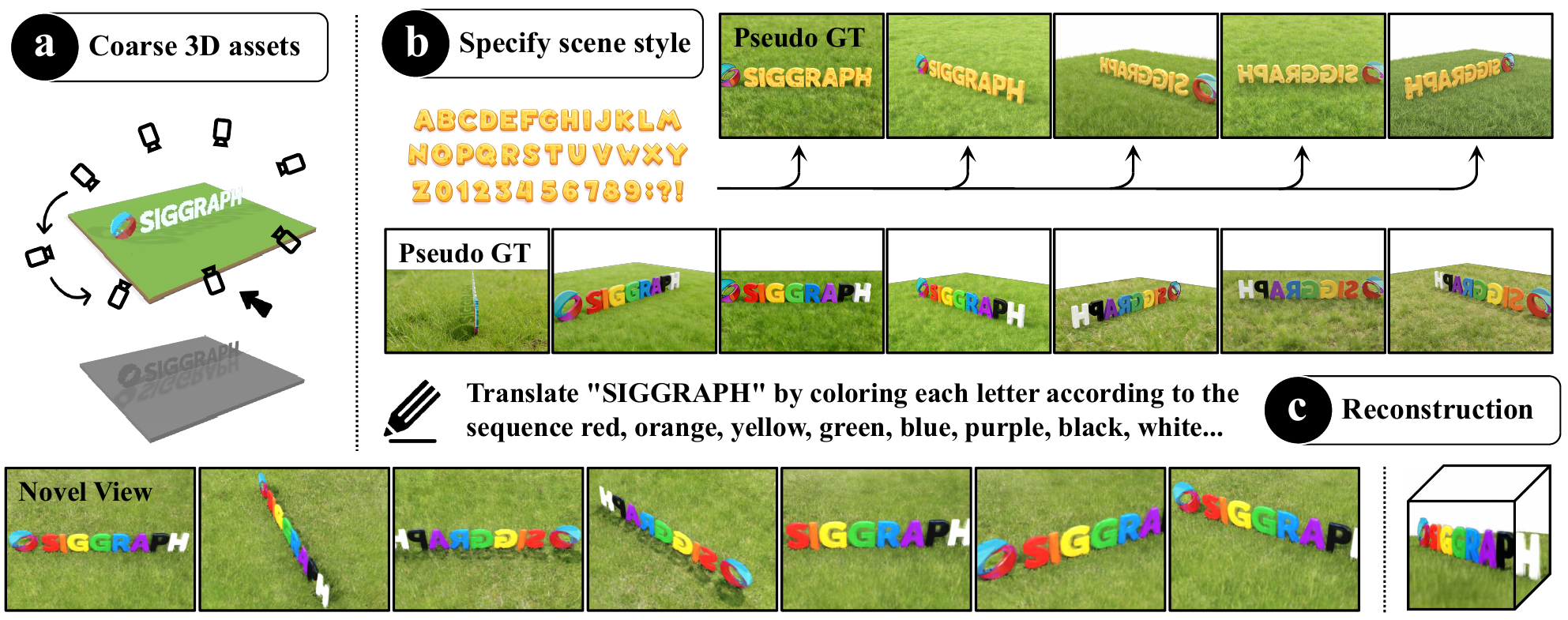}
  \vspace{-6mm}
  \caption{Overview framework. (a) Users first initialize a coarse geometric scene using manual assembly (e.g., Minecraft) or existing 3D meshes. (b) By specifying the desired visual style via reference images or text prompts, the model generates multi-view pseudo-GTs as supervision. (c) Finally, our method reconstructs a high-quality 3D scene that maintains both style consistency and structural regularity.}
  \label{fig:teaser}
\end{teaserfigure}

\title{GeoDiff3D: Self-Supervised 3D Scene Generation with Geometry-Constrained 2D Diffusion Guidance}

\author{Haozhi Zhu}
\affiliation{%
  \institution{Nanjing University}
  \city{Nanjing}
  \postcode{210023}
  \country{China}
}
\email{662024330006@smail.nju.edu.cn}

\author{Miaomiao Zhao}
\authornote{Equal contribution.}
\affiliation{%
  \institution{Nanjing University}
  \city{Nanjing}
  \postcode{210023}
  \country{China}}
\email{522025330152@smail.nju.edu.cn}

\author{Dingyao Liu}
\affiliation{%
  \institution{Nanjing University}
  \city{Nanjing}
  \postcode{210023}
  \country{China}
}
\email{522024330044@smail.nju.edu.cn}

\author{Runze Tian}
\affiliation{%
  \institution{Nanjing University}
  \city{Nanjing}
  \postcode{210023}
  \country{China}
}
\email{522024330075@smail.nju.edu.cn}

\author{Yan Zhang}
\authornote{Corresponding author. Email: zhangyannju@nju.edu.cn}
\affiliation{%
  \institution{Nanjing University}
  \city{Nanjing}
  \postcode{210023}
  \country{China}
}

\author{Jie Guo}
\authornote{Corresponding author. Email: guojie@nju.edu.cn}
\affiliation{%
  \institution{Nanjing University}
  \city{Nanjing}
  \postcode{210023}
  \country{China}
}

\author{Fenggen Yu}
\affiliation{%
  \institution{Simon Fraser university}
  \country{China}
}
\email{yufg1994@gmail.com}

\begin{abstract}
3D scene generation is a core technology for gaming, film/VFX, and VR/AR. Growing demand for rapid iteration, high-fidelity detail, and accessible content creation has further increased interest in this area. Existing methods broadly follow two paradigms—indirect 2D-to-3D reconstruction and direct 3D generation—but both are limited by weak structural modeling and heavy reliance on large-scale ground-truth supervision, often producing structural artifacts, geometric inconsistencies, and degraded high-frequency details in complex scenes. We propose GeoDiff3D, an efficient self-supervised framework that uses coarse geometry as a structural anchor and a geometry-constrained 2D diffusion model to provide texture-rich reference images. Importantly, GeoDiff3D does not require strict multi-view consistency of the diffusion-generated references and remains robust to the resulting noisy, inconsistent guidance. We further introduce voxel-aligned 3D feature aggregation and dual self-supervision to maintain scene coherence and fine details while substantially reducing dependence on labeled data. GeoDiff3D also trains with low computational cost and enables fast, high-quality 3D scene generation. Extensive experiments on challenging scenes show improved generalization and generation quality over existing baselines, offering a practical solution for accessible and efficient 3D scene construction.
\end{abstract}

%
%


%
%


\maketitle
  
\section{Introduction}
\label{sec1:introduction}

3D scene generation, as a core topic spanning computer graphics, artificial intelligence, and virtual world construction, has garnered significant attention.
From open-world creation in games and virtual set construction in films to immersive environments in VR applications, the demand for high-quality 3D scenes continues to grow, with increasing emphasis on rapid iteration, rich visual details, and low barriers to content creation.

In recent years, diffusion-based generative models have achieved remarkable success in image generation ~\cite{Rombach_2022_CVPR, NEURIPS2020_4c5bcfec, NEURIPS2022_ec795aea} and video generation~\cite{yang2024cogvideox, guo2023animatediff, blattmann2023stablevideodiffusionscaling}, offering flexible and efficient solutions for visual content creation.
Inspired by this progress, plenty of methods~\cite{liu2024reconxreconstructscenesparse, Long_2024_CVPR, instant3d2023} have explored reconstructing 3D scenes from multi-view images or videos generated by 2D diffusion models.
However, 2D diffusion models excel at producing visual realistic 2D content, but lack explicit constraints and global consistency of 3D structure.
In contrast, 3D scene generation requires strict structural coherence and visual consistency.
This fundamental mismatch often causes structural artifacts, geometric inconsistencies, and violations of physical constraints in 2D-based reconstruction methods.

Besides, researchers have extended the idea of diffusion models to 3D~\cite{poole2022dreamfusiontextto3dusing2d, Lin_2023_CVPR, Tang_2023_ICCV}. 
These methods learn 3D generative priors from large-scale 3D datasets and can produce complete 3D outputs from text prompts, sparse views, and even single images.
While effective for simple object generation, most existing methods~\cite{xiang2024structured, xiang2025trellis2, chang2025reconviagen} struggle to generate complex scenes such as detailed building layouts or large-scale natural environments.
This limitation arises from the lack of high quality 3D training data and the complexity of real-world scenes, leading to disorganized structures and blurred details.
In summary, whether following an indirect 2D-to-3D reconstruction paradigm or a direct 3D generation paradigm, current methods remain constrained by insufficient modeling of 3D structural priors and a heavy reliance on large training datasets.

Reviewing traditional 3D scene generation pipelines reveals a key insight for addressing these challenges: coarse geometric structures serve as the foundational anchor through the entire modeling process.
In conventional workflows, artists first construct rough geometric layouts and then progressively refine textures, lighting, and materials through labor-intensive procedures.
Although this process demands substantial expertise and limits rapid iteration, coarse geometry enforces precise spatial constraints, ensuring correct proportions, structural validity, and view consistency.
Such geometric anchoring effectively compensates for the lack of explicit 3D structural constraints in existing generative methods and remains indispensable for reliable scene construction.

Meanwhile, despite the difficulty of directly reconstructing complete 3D models from 2D diffusion models, these methods~\cite{liu2024reconxreconstructscenesparse, Long_2024_CVPR, instant3d2023} remain highly complementary to coarse geometry.
On one hand, 2D diffusion models excel at generating high-quality images with fine-grained local details, which can serve as high-fidelity texture sources without requiring large-scale 3D datasets.
On the other hand, their flexibility in visual concept exploration enables efficient style variation and rapid design iteration.
Therefore, 2D diffusion models can effectively bridge coarse geometric structures and high-quality 3D scene generation, balancing structural correctness and visual quality.

Based on these observations, we propose a novel framework for high-quality 3D scene generation that integrates coarse geometric constraints with the strengths of 2D diffusion models.
Our method avoids complex manual post-processing and heavy reliance on large annotated datasets.
By using coarse geometry as a structural anchor and leveraging diffusion-based image generation for detail synthesis, we introduce a voxel-aligned generative modeling framework that automates texture generation and detail completion.
This framework significantly simplifies the 3D scene creation pipeline, offering an efficient and low-barrier alternative to traditional labor-intensive workflows.
~\cref{fig:teaser} illustrates the overall pipeline.
As shown in ~\cref{fig:teaser}(a), users first construct a coarse geometric scene through manual modeling or asset assembly.
Then, as shown in ~\cref{fig:teaser}(b), reference images or text prompts are provided to specify the desired visual style.
Finally, as shown in ~\cref{fig:teaser}(c), our method generates high-quality 3D scenes with both structural coherence and consistent visual appearance.

Specifically, our framework consists of a three-stage progressive pipeline that unifies geometric constraints and realistic details.
First, we generate texture reference images.
Based on the coarse geometry and the target style, we use 2D diffusion models~\cite{zhang2023adding, mou2023t2i, kim2025videofrom3d} to generate texture-rich yet multi-view inconsistent images, which serve as pseudo-ground-truth (pseudo-GT).
Second, we perform 3D feature aggregation and optimization.
Inspired by voxel-aligned representations ~\cite{wang2025volsplat, xiang2024structured, jiang2025anysplat}, features extracted from the pseudo-GT images are aligned and aggregated into voxel grids corresponding to the coarse geometry.
A sparse 3D decoder then refines the aggregated features and predicts voxel-aligned Gaussian distributions.
This volumetric aggregation effectively mitigates floating artifacts and view inconsistency, transferring multi-view information to shared 3D representations before detail synthesis.
Third, we introduce a self-supervised training strategy to address the potential multi-view inconsistency of the pseudo-GT.
We jointly leverage diffusion-generated reference images, which preserve visual realism, and novel-view renderings from the voxel model, which enforce structural consistency.
By designing a tailored set of objective functions, our method effectively balances structural coherence and visual fidelity while significantly reducing reliance on large-scale datasets.
Our main contributions include:
\begin{itemize}
\item We propose ~\textit{GeoDiff3D}, a geometry-constrained 2D diffusion–assisted self-supervised framework for 3D scene generation, enabling efficient and high-quality 3D scene creation while effectively addressing the core limitations of existing methods, namely weak structural constraint and heavy dependence on large-scale dataset.
\item We introduce a voxel-aligned 3D feature aggregation mechanism that jointly preserves generation quality and structural coherence, reduces over-reliance on reference image consistency, and ensures the completeness of the generated structure.
\item We design a dual self-supervised optimization strategy that precisely balances structural consistency and realistic details while significantly reducing reliance on large-scale annotated dataset.
\item Our method does not require high-end computational resources and can rapidly generate high-quality 3D scenes, making it well suited for low-barrier, high-efficiency 3D content creation.
\end{itemize}

\section{Related Work}
\label{sec2:relatedwork}

\indent \indent \textit{Reconstruction-based 3D Model Creation.}
In recent years, a class of methods~\cite{yu2024viewcrafter, gao2024cat3d, jiang2025anysplat, Nam_2025_ICCV} for building 3D scenes from scratch first generates multi-view images or videos using diffusion models, followed by 3D reconstruction using techniques such as 3D Gaussian Splatting (3DGS)~\cite{kerbl3Dgaussians} or NeRF~\cite{mildenhall2020nerf}.
ReconX~\cite{liu2024reconxreconstructscenesparse} uses video diffusion models to synthesis video frames as dense input to guide 3DGS optimization.
AnySplat~\cite{jiang2025anysplat} leverages camera and depth priors learned from large-scale datasets and directly acquires 3DGS parameters from image features via a feed-forward network. 
Voyager~\cite{huang2025voyager} uses depth maps and camera trajectories as constraints to guide the generative model to synthesize videos with explicit spatial logic, further enabling large-scale scene construction through a world cache. 
Marble~\cite{MarbleWorldLabs} generates an explorable 3D world from a single image input.
However, these methods only optimize in the 2D pixel space and lack explicit 3D geometric guidance, leading to poor multi-view consistency and a disconnect between generation and reconstruction. 
In contrast, our method uses coarse geometry as a core structural constraint and tightly combines generation and reconstruction through voxel-aligned feature aggregation and a dual self-supervised optimization strategy, ensuring structural consistency while reducing drift and artifacts.

\indent \indent \textit{Generation-based 3D Model Creation.}
Leveraging the strong priors of generative models for 3D reconstruction~\cite{cao2025uni3c, Wu_2024_CVPR, liu20243dgs, chen2024mvsplat360} has become a major research direction.
See3D~\cite{Ma2025See3D} is the first to train a 3D generative model directly on large-scale in-the-wild videos without pose annotations, and further supports editing. 
RomanTex~\cite{Feng_2025_ICCV} injects geometric structure into the attention mechanism of a multi-view diffusion model, improving cross-view texture consistency.
ReconViaGen~\cite{chang2025reconviagen} combines reconstruction priors from VGGT~\cite{Wang_2025_CVPR} with diffusion generation priors to enhance controllability. 
InstantMesh~\cite{xu2024instantmesh} combines multi-view diffusion with a sparse reconstruction network to rapidly produce topologically consistent meshes.
Despite their impressive progress, most of these methods heavily rely on large-scale annotated 3D datasets and are limited to single-object generation, exhibiting structural degradation and poor generalization in complex scenes.
Our method leverages the rich detail generation capability of 2D diffusion models to enhance scene textures, and significantly improves generalization in complex scenes through a dual self-supervised strategy, without requiring massive labeled 3D datasets.

\indent \indent \textit{Voxel Representations.}
Voxel grids, with their regular structure, serve as an effective bridge between 2D visual observations and explicit 3D geometric representations.
To address feature alignment, VolSplat~\cite{wang2025volsplat} constructs a feature-matching cost volume and refines it with 3D U-Net.
To improve rendering efficiency, AnySplat~\cite{jiang2025anysplat} introduces Gaussian voxelization, aggregating and filtering Gaussian primitives within voxel grids to significantly reduce redundant computation. 
UniSplat~\cite{shi2025unisplat} builds a 3D latent scaffold as structural support, enabling flexible capture of time-varying geometry and achieving high-quality dynamic reconstruction for more complex scenes. 
Trellis 1.0~\cite{xiang2024structured} introduces a structured latent variable model that compresses complex 3D data into sparse voxel latents and trains a decoder to recover high-fidelity geometry and textures.
These works demonstrate the effectiveness of voxel representations for 3D reconstruction.
Our method leverages voxels for feature fusion, aligning detailed features generated by 2D diffusion models into 3D space to form a unified representation, balancing structural constraints and visual fidelity.
\section{Method}
\label{sec3:method}

\begin{figure*}[t]
  \centering
  \includegraphics[width=1\linewidth]{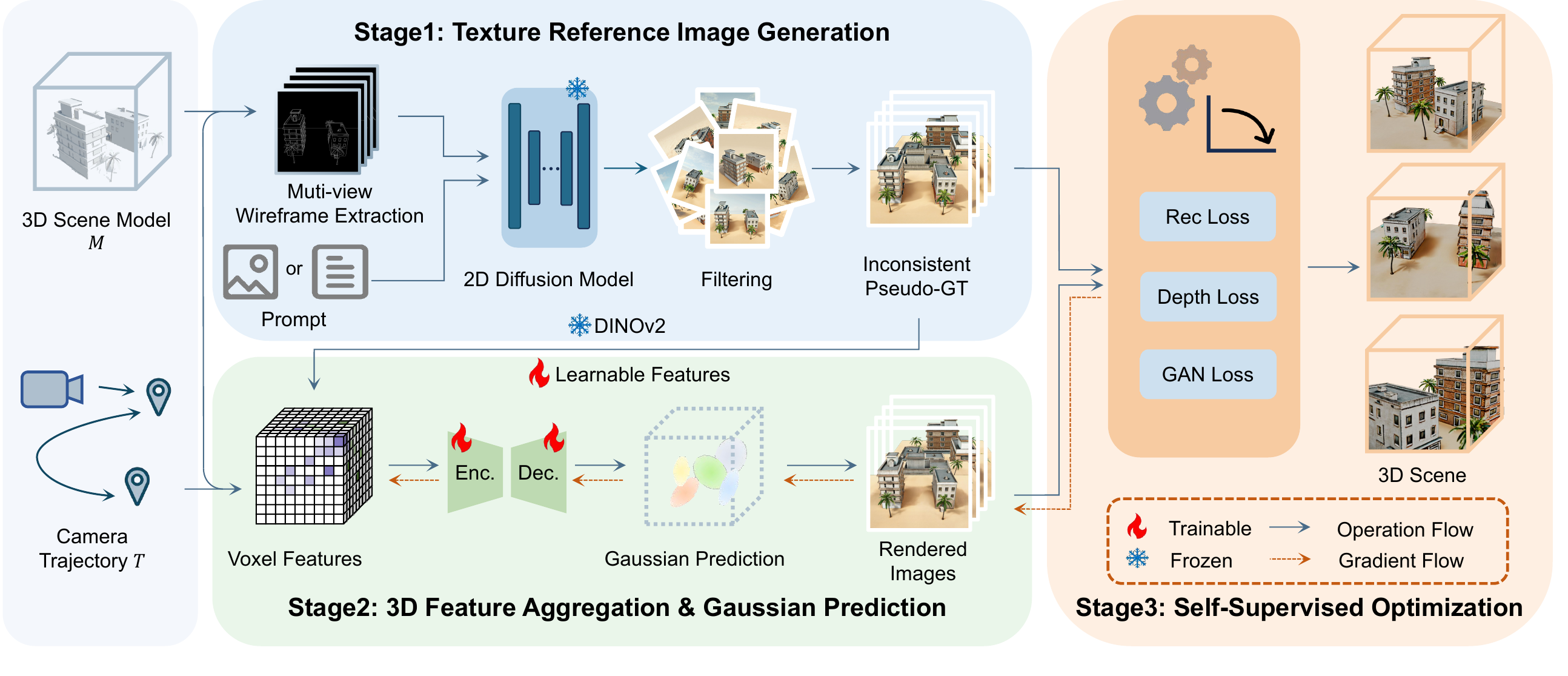}
  \vspace{-6mm}
   \caption{Overview of GeoDiff3D. In the first stage, we extract structural edges from the input 3D model along the camera trajectory and generate multi-view pseudo-GT images guided by an image diffusion prior. In the second stage, we back-project 2D features from the selected pseudo-GT views into 3D space to form a voxel feature volume, which is then decoded into a 3D Gaussian (3DGS) representation. In the third stage, we perform self-supervised optimization of the generated 3D scene using reconstruction loss, depth loss, and GAN loss.}
   \label{fig:overview}
\end{figure*}

We define the task of ~\textit{GeoDiff3D} as follows: let $M$ denote a 3D scene model which supports multiple representations such as meshes and voxels, and let a reference image $I_{ref}$ or a text prompt $T$ specify the target visual style. Given these inputs, our goal is to generate a 3D scene represented as 3D Gaussian Splatting (3DGS)~\cite{kerbl3Dgaussians} that faithfully preserves the geometry of $M$ while achieving visual consistency with the style of $I_{\mathrm{ref}}$ or $T$.

As illustrated in ~\cref{fig:overview}, GeoDiff3D consists of three stages.
First, in the texture reference generation stage, we take the 3D scene model $M$ and camera trajectories as input, extract multi-view projected edge constraints as structural guidance, and combine them with the reference style.
Then we use a 2D diffusion model~\cite{kim2025videofrom3d} to synthesize pseudo–ground-truth (pseudo-GT) texture reference images as guides for the scene generation.
Next, in the 3D feature aggregation and optimization stage, the pseudo-GT image features are aligned and aggregated into the voxel grids corresponding to $M$.
A sparse 3D decoder refines these features and predicts voxel-aligned Gaussian distributions to ensure cross-view consistency.
Finally, we perform self-supervised optimization using the pseudo-GT images and the rendered images decoded from the second stage, generating a high-quality 3D scene model.

\subsection{Texture Reference Image Generation}
To achieve high-quality 3D scene reconstruction, we first construct pseudo-GT texture reference images that preserve the scene structure while providing rich style details.
This stage combines two complementary cues: the coarse 3D model provides geometric guidance (e.g., contours and depth ordering), while a reference image or text prompt specifies the target style. Leveraging these cues, we use a 2D diffusion model to generate multi-view images.
We then select high-quality results as pseudo-GT, which form the foundation for subsequent 3D feature aggregation and scene optimization.

\textit{Geometry-Guided Texture Generation.}
Given predefined camera parameters, the coarse 3D model is first projected into a set of textureless 2D images containing only depth and contour information. Inspired by the anchor-based strategy of VideoFrom3D~\cite{kim2025videofrom3d}, We then employ Flux-ControlNet~\cite{XLabs2024} to inject the extracted line maps as structural priors into the 2D diffusion process, enabling controllable stylized texture synthesis. A reference image or text prompt further specifies the target appearance, including texture characteristics and artistic style. As a result, we obtain multi-view pseudo-GT images with rich textures that are consistent with the reference style.

\textit{Multi-Dimensional Quality Filtering Strategy.}
Although 2D diffusion models bring vivid details, their inherent randomness introduces notable limitations like content hallucination and multi-view inconsistency.
To ensure the reliability of pseudo-GT, we design a lightweight filtering strategy to prune the generated images.

First, to prevent off-topic generation, we use CLIP~\cite{radford2021learningtransferablevisualmodels} to measure the semantic similarity between the generated images and the input reference image or text description and discard low-scoring samples, ensuring that each pseudo-GT faithfully reflects the intended scene semantics and style. Then, we mask the generated images using the textureless 2D projections rendered from the coarse 3D model to filter samples with geometric distortions. Finally, we select a varying number of pseudo-GTs as input based on the size of the scene.

\subsection {3D Feature Aggregation and Gaussian Prediction}
\label{sec3-2}
After obtaining $V$ filtered pseudo-GT images, we could directly apply any 3DGS-based method~\cite{ren2024octree, zhu2023FSGS, zhang2024pixelgs} to reconstruct the scene. However, this suffers from two key limitations:
(1) the pseudo-GT generation stage does not explicitly enforce cross-view consistency, leading to geometric distortions such as floating artifacts in 3DGS reconstruction;
(2) 3DGS requires a large number of input images, and generating more pseudo-GT images accumulates cross-view inconsistency, further degrading reconstruction quality.

Inspired by voxel-based 3D reconstruction methods~\cite{xiang2024structured, wang2025volsplat, jiang2025anysplat}, we take the $v$ pseudo-GT images and their camera poses as input, and learn to align and aggregate them into a voxel representation of the coarse geometry. 
A sparse 3D decoder then refines the 3D features and predicts voxel-aligned Gaussian distributions. 

\textit{Feature Extraction and Matching.}
Given a scene model $M$, the pseudo-GT image sequence 
$\left\{ I_v \right\}_{v=1}^{V}$ rendered from multiple viewpoints, and the corresponding camera parameters, we first voxelize the scene into a fixed-resolution sparse 3D voxel grid and extract the occupied voxel set with their indices.
For each view $I_v$, we extract patch-level ViT features $F_v$ using a pretrained DINOv2~\cite{oquab2024dinov2learningrobustvisual} encoder. 
To align 2D semantic information with the 3D voxel space, we project each occupied voxel into each view using the camera intrinsics and extrinsics, and bilinearly sample the corresponding feature from $F_v$.
Finally, we average the voxel features across views to improve robustness to cross-view inconsistency and view-dependent noise, acquiring semantically enriched voxel features.

\textit{Feature Refinement.}
The aggregated voxel feature $f_p$ mainly encodes globally consistent semantics across views. While robust, average aggregating smooths out local high-frequency details, making it insufficient for decoding fine-grained geometry and texture.
To address this limitation, we introduce a learnable feature residual $\Delta f_p \in \mathbb{R}^d$ for each occupied voxel and define the refined voxel feature as:
\begin{equation}
\label{eqn:feat_refine}
\tilde{f}_p = f_p + \Delta f_p,
\end{equation}
where $f_p$ is the aggregated feature. $\Delta f_p$ is optimized jointly with the network parameters to adaptively compensate for information lost during aggregation. This design preserves global consistency while restoring local details, enabling accurate and detailed Gaussian parameter decoding in subsequent stages.

\textit{Gaussian Prediction.}
Inspired by the Trellis 1.0~\cite{xiang2024structured} encoder–decoder framework, we employ a sparse 3D VAE to map the refined voxel features $\tilde{f}_p$ to a renderable Gaussian representation.
For each voxel $p$, the parameters of its $j-th$ Gaussian are defined as 
\begin{equation}
\mathbf{g}_{p,j}=\big(\mathbf{o}_{p,j},\ \bar{\alpha}_{p,j},\ \mathbf{s}_{p,j},\ \mathbf{q}_{p,j},\ \mathbf{c}_{p,j}\big), j=1,\dots,32 
\end{equation}
where the parameters correspond to the center offset, opacity, scaling, rotation and color, respectively.

To ensure that the predicted parameters lie within valid ranges, we apply bounded mappings: the opacity is constrained to $(0,1)$ with a Sigmoid function, and the Gaussian center is determined from the voxel center $\mathbf{x}_p$ and the constrained offset. Specifically, 
set the voxel length as $\ell$, the Gaussian center is computed as:
\begin{equation}
\label{eqn:gauss_center_mapping}
\boldsymbol{\mu}_{p,j} = \mathbf{x}_p + \ell \cdot \Big(\sigma(\mathbf{o}_{p,j}) - \tfrac{1}{2}\Big),
\qquad
\alpha_{p,j} = \sigma(\bar{\alpha}_{p,j}),
\end{equation}
where $\mathbf{o}_{p,j}\in\mathbb{R}^3$ is the center offset predicted by the decoder, and $\sigma(\cdot)$ denotes the Sigmoid function.
This parameterization anchors each Gaussian within a local neighborhood aligned with the voxel scale, improving training stability and enforcing alignment with the voxel grid structure.
 
\textit{Deterministic Template Perturbation.}
With voxel-anchored parameterization, neighboring voxels tend to produce similar offset patterns due to local smoothness. During optimization, Gaussian centers may also be pushed toward the boundaries of the offset range, leading to grid-aligned blocky artifacts in rendering.
To alleviate this regularity, we add a deterministic voxel-dependent perturbation to the Gaussian centers:
\begin{equation}
\label{eqn:gauss_center_refine}
\boldsymbol{\mu}_{p,j} \leftarrow \boldsymbol{\mu}_{p,j} + \rho \cdot \ell \cdot \mathbf{t}_{p,j},
\end{equation}
where $\mathbf{t}_{p,j}\in[-1,1]^3$ is determined from the voxel index and Gaussian number, 
$\rho$ denotes the perturbation radius.
This perturbation weakens strict grid alignment while preserving the global geometry, ensuring more continuous and visually stable renderings.

\subsection{Self-Supervised Optimization}
After the pipeline described in ~\cref{sec3-2}, voxel-rendered images derived from the coarse geometry preserve spatial coherence. However, they are limited by the cross-view inconsistency of pseudo-GT, often leading to blurred details and insufficient realism.
Existing voxel-based reconstruction methods~\cite{wang2025volsplat, jiang2025anysplat} rely on large-scale annotated 3D datasets for supervised training, which is impractical in the from-scratch reconstruction where such annotations are unavailable.

Inspired by MV2MV~\cite{Cai2024MV2MV}, which constructs supervision from complementary pseudo labels, we design a dual-supervision self-supervised optimization strategy.
Pseudo-GT from the 2D diffusion model provides realistic visual details, while novel views rendered from the voxel model provide complementary structural consistency.
Consistency loss enhances multi-view coherence, and GAN loss preserves high-fidelity visual details.
Within this self-supervised framework, we jointly optimize the encoder–decoder network and the learnable voxel feature residuals ($\Delta f_p$ in ~\cref{sec3-2}) in an end-to-end manner.
This strategy eliminates the dependence on large-scale annotated 3D data and effectively bridges the gap between 2D diffusion generation and 3D structural requirements, enabling robust and high-quality 3D scene generation. The pipeline is defined as follows:

\textit{Consistency Loss.}
We adopt the standard 3DGS reconstruction loss to supervise the optimization of Gaussians.
We render the current Gaussian set to obtain $\hat{I}_v$ for each view, and calculate a weighted combination of pixel-wise $L_1$, $D-SSIM$ and $LPIPS$ losses losses with the corresponding pseudo-GT image $I_v$.
This reconstruction loss suppresses view-dependent noise and appearance drift, thereby improving cross-view consistency and rendering stability. We define the reconstruction loss as follows:
\begin{equation}
\label{eq:reconstruction_loss}
\mathcal{L}_{\mathrm{rec}} = \lambda_{\mathrm{L1}} \cdot \mathcal{L}_{\mathrm{L1}} + \lambda_{\mathrm{D-SSIM}} \cdot \mathcal{L}_{\mathrm{D-SSIM}} + \lambda_{\mathrm{LPIPS}} \cdot \mathcal{L}_{\mathrm{LPIPS}}
\end{equation}

To further enhance geometric consistency, we introduce patch-based depth regularization~\cite{li2024dngaussian} on the depth map $d$.
Specifically, we divide the depth map into separate local patches $P$, and normalize depth both locally and globally to constrain local depth variations and global shape trends.

\begin{equation}
\label{eq:depth_ln}
d_{\mathrm{LN}}(x)=\frac{d(x)-\mathrm{mean}(d(P))}{\mathrm{std}(d(P))+\epsilon},
\end{equation}
\begin{equation}
\label{eq:depth_gn}
d_{\mathrm{GN}}(x)=\frac{d(x)-\mathrm{mean}(d(P))}{\mathrm{std}(d(I))+\epsilon},
\end{equation}
where $x\in P$, and $I$ denotes the entire image field. We use $\hat d$ for the rendered depth and $d^{*}$ for the ground-truth depth.

In practice, to avoid numerical instability caused by extremely small variances, we add a stability term $\epsilon$ related to the global variance to the denominator.
In addition, we use a truncated $L_2$ norm with a tolerance threshold $\tau$ to reduce the influence of noise supervision, where errors are only penalized when they exceed the threshold: 
\begin{equation}
\label{eq:trunc_l2}
\ell_{\tau}(a,b)=\mathbb{I}(|a-b|>\tau)\cdot (a-b)^2 .
\end{equation}
The final depth regularization term is a weighted combination of the local and global term:
\begin{equation}
\label{eq:depth_loss}
\mathcal{L}_{\mathrm{depth}}
=\lambda_{\mathrm{L}}\;\mathbb{E}_{P}\big[\ell_{\tau}(\hat{d}_{\mathrm{LN}},d^{*}_{\mathrm{LN}})\big]
+\lambda_{\mathrm{G}}\;\mathbb{E}_{P}\big[\ell_{\tau}(\hat{d}_{\mathrm{GN}},d^{*}_{\mathrm{GN}})\big].
\end{equation}

\textit{GAN Loss.}
Since pseudo-GT generated by 2D diffusion model inevitably exhibits cross-view inconsistency, relying solely on consistency losses tends to average appearances across views, resulting in over-smoothed and blurry textures.
To address this, we introduce adversarial training with a PatchGAN~\cite{Isola_2017_CVPR} discriminator $D(\cdot)$, which enables the model to learn discriminative high-frequency details from pseudo-GT while maintaining global structural consistency.

The discriminator outputs a spatial score map $D(I)\in\mathbb{R}^{H'\times W'}$ for an input image and its mean value is denoted as $\overline{D(I)}$.
During adversarial training, high-quality pseudo-GT images generated by the 2D diffusion model are treated as real samples $I\sim p_{\mathrm{data}}$, while images rendered from the current 3D representation are treated as fake samples $\hat I\sim p_G$.
We adopt a hinge style adversarial loss, and the discriminator target is defined as:
\begin{equation}
\label{eqn:gan_discriminator}
\begin{split}
\mathcal{L}_{D}^{\mathrm{GAN}}
= \frac{1}{2}\Big(
\mathbb{E}_{I\sim p_{\mathrm{data}}}\big[\ \overline{\max(0,\,1 - D(I))}\ \big]
+ {}\\
\mathbb{E}_{\hat I\sim p_G}\big[\ \overline{\max(0,\,1 + D(\hat I))}\ \big]
\Big).
\end{split}
\end{equation}

The generator, including encoder-decoder and the rendering module, has an GAN loss defined as:
\begin{equation}
\label{eqn:gan_generator}
\mathcal{L}_{G}^{\mathrm{GAN}}
=
-\,\mathbb{E}_{\hat I\sim p_G}\big[\ \overline{D(\hat I)}\ \big],
\end{equation}

We incorporate the GAN and depth losses into the total generator objective with weights $\lambda_{\mathrm{gan}}$ and $\lambda_{\mathrm{depth}}$, respectively:
\begin{equation}
\label{eqn:gen_total_loss}
\mathcal{L}_G = \mathcal{L}_{\mathrm{rec}} + \lambda_{\mathrm{depth}} \cdot \mathcal{L}_{\mathrm{depth}} + \lambda_{\mathrm{gan}} \cdot \mathcal{L}_G^{\mathrm{GAN}}
\end{equation}

\section{Experiments}
\label{sec4:experiments}

\subsection{Implementation details}
In the first-stage of pseudo-GT generation, we generate all pseudo-GT images at a fixed resolution of $1024 \times 1024$. Depending on the scene scale, we use different numbers of pseudo-GT views: $10-12$ views for large-scale scenes and $4-8$ views for smaller scenes. 
During training, we set the voxel grid resolution to $128^3$.
The weights of reconstruction-related losses are set following the original 3DGS framework, with the depth loss weight set to $0.1$.
The GAN loss is assigned a weight of 0.05 and is introduced only after 500 training iterations to avoid disrupting early geometric structure learning, by which time the scene typically already has a relatively stable 3D structure.
We employ two separate optimizers for different objectives: one for consistency-related losses with a learning rate of 
$1 \times 10^{-4}$, and another for the GAN loss with a learning rate of $5 \times 10^{-6}$. 
Each scene is trained for $1,000$ iterations on a single NVIDIA A800 (40GB) GPU, with a total training time of approximately ten to twenty minutes.

\subsection{Model Generation results}
\begin{figure*}[t]
  \includegraphics[width=0.9 \textwidth, keepaspectratio]{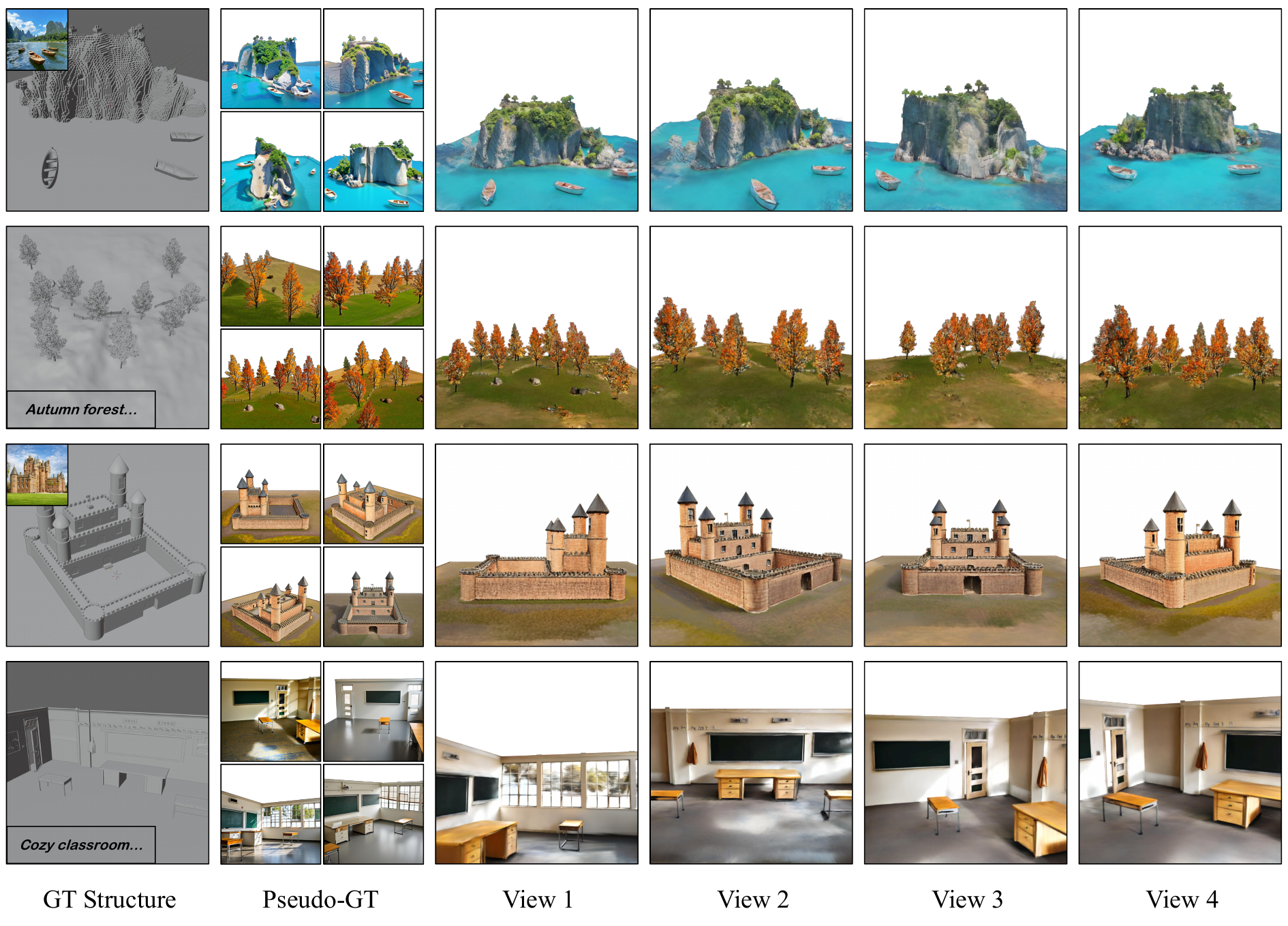}
  \caption{Qualitative results across various scenes. The first column presents the inputs, including the style reference (top-left) and the text prompt (bottom-left). The second column shows representative pseudo-GT images generated by our method. The remaining images show our renderings from different viewpoints.}
  \label{fig:result}
\end{figure*}

\begin{figure*}
  \includegraphics[width=\textwidth, keepaspectratio]{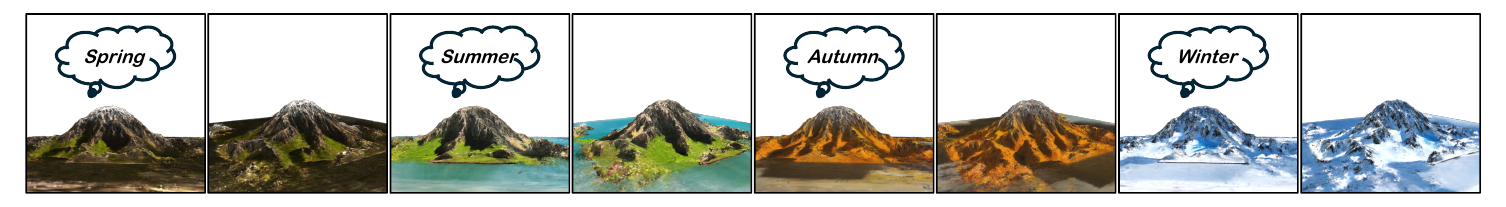}
  \caption{Seasonal style transfer on the same scene. Given different style references, our method can switch the scene's appearance (e.g., four seasons) while preserving its geometry.}
  \label{fig:season}
\end{figure*}

~\cref{fig:result} presents qualitative results of our method across diverse scenes.
We consider a variety of inputs, including Minecraft-style scenes and untextured scenes.
The results in the first two rows demonstrate strong robustness for large-scale outdoor scenes: even with complex spatial layouts and rich environmental details, our method achieves high visual quality with well-preserved fine details.
The last two rows highlight consistently stable performance on buildings-centric scenes, where complex geometric structures such as fine contours and multi-component assemblies are faithfully reconstructed, generating high-fidelity scene models that accurately capture architectural characteristics. 
These results clearly demonstrate the adaptability of our method across different scene types.
In addition, ~\cref{fig:season} illustrates the fine-grained style controllability of our method with different modifier prompts, further validating its flexibility, efficiency, and scalability for style control.
Additional results are provided in the supplementary material.

\subsection{Baseline Comparisons}

\indent \indent \textit{Qualitative Comparisons.}
To assess the efficacy of our method, we conduct comparative experiments against five representative methods, including the image-based generative method Trellis 1.0~\cite{xiang2024structured}, the world models World-Mirror~\cite{liu2025worldmirror} and Marble~\cite{MarbleWorldLabs}, the 3D geometry constrained video synthesis method VideoFrom3D~\cite{kim2025videofrom3d}, and the sparse-view 3DGS reconstruction method FSGS~\cite{zhu2023FSGS}.

\begin{table}[!t]
\caption{Quantitative comparisons with Trellis 1.0~\cite{xiang2024structured}, World-Mirror~\cite{liu2025worldmirror}, FSGS~\cite{zhu2023FSGS} and VF3D~\cite{kim2025videofrom3d}.}
\vspace{-3mm}
\belowrulesep=0pt
\aboverulesep=0pt
\centering
\renewcommand\arraystretch{1.3}
\setlength{\tabcolsep}{3pt} 
\begin{center}
\begin{tabular}{c|ccccc}
\toprule
Method & PSNR-D $\uparrow$ & MUSIQ $\uparrow$ & MANIQ $\uparrow$ & CC $\uparrow$ & CS $\uparrow$ \\
\midrule
Trellis 1.0 & 17.41 & 44.71 & 0.26 & 0.92 & 0.81 \\
World-Mirror & 15.09 & 50.76 & 0.25 & 0.90 & 0.76 \\
FSGS & 18.26 & 57.07 & 0.31 & 0.90 & 0.84 \\
VF3D+3DGS & 17.46 & 48.73 & 0.27 & 0.85 & 0.86 \\
Ours & \textbf{20.39} & \textbf{58.06} & \textbf{0.35} & \textbf{0.93} & \textbf{0.90} \\
\bottomrule
\end{tabular}
\end{center}
\label{tab:table1}
\end{table}

\begin{table}[!t]
\belowrulesep=0pt
\aboverulesep=0pt
\centering
\renewcommand\arraystretch{1.3}
\caption{Quantitative comparisons with Marble~\cite{MarbleWorldLabs}.}
\vspace{-3mm}
\setlength{\tabcolsep}{3pt} 
\begin{tabular}{c|ccccc}
\toprule
Method & PSNR-D $\uparrow$ & MUSIQ $\uparrow$ & MANIQ $\uparrow$ & CC $\uparrow$ & CS $\uparrow$ \\
\midrule
Marble & 16.68 & 55.75 & \textbf{0.35} & 0.90 & 0.82 \\
Ours & \textbf{20.39} & \textbf{58.06} & \textbf{0.35} & \textbf{0.93} & \textbf{0.90} \\
\bottomrule
\end{tabular}
\label{tab:table2}
\end{table}

The experimental settings are as follows. For Trellis 1.0, we treat the existing sparse voxel representation as the sparse structural prediction from its first stage and feed multi-view pseudo-GT images into its second-stage generation module.
For World-Mirror, we provide camera poses and ground-truth depth as prior information. 
For Marble in indoor scenes, we import an existing indoor model into its indoor pipeline and perform 3D generation conditioned on style descriptions derived from reference images. 
Since Marble relies on panoramic inputs for outdoor scenes and is not applicable to the inconsistent pseudo-GT setting in our framework, we report comparisons with Marble only on indoor scenes.
For VideoFrom3D, We incorporate our pseudo-GT as anchor points in the video generation and reconstruct 3D scenes using 3DGS based on its synthesized video outputs.
For FSGS, we use the pseudo-GT images from our first stage as input, and initialize the COLMAP point cloud with our densified sparse voxel points.

\begin{figure*}
  \includegraphics[width=0.8 \textwidth, keepaspectratio]{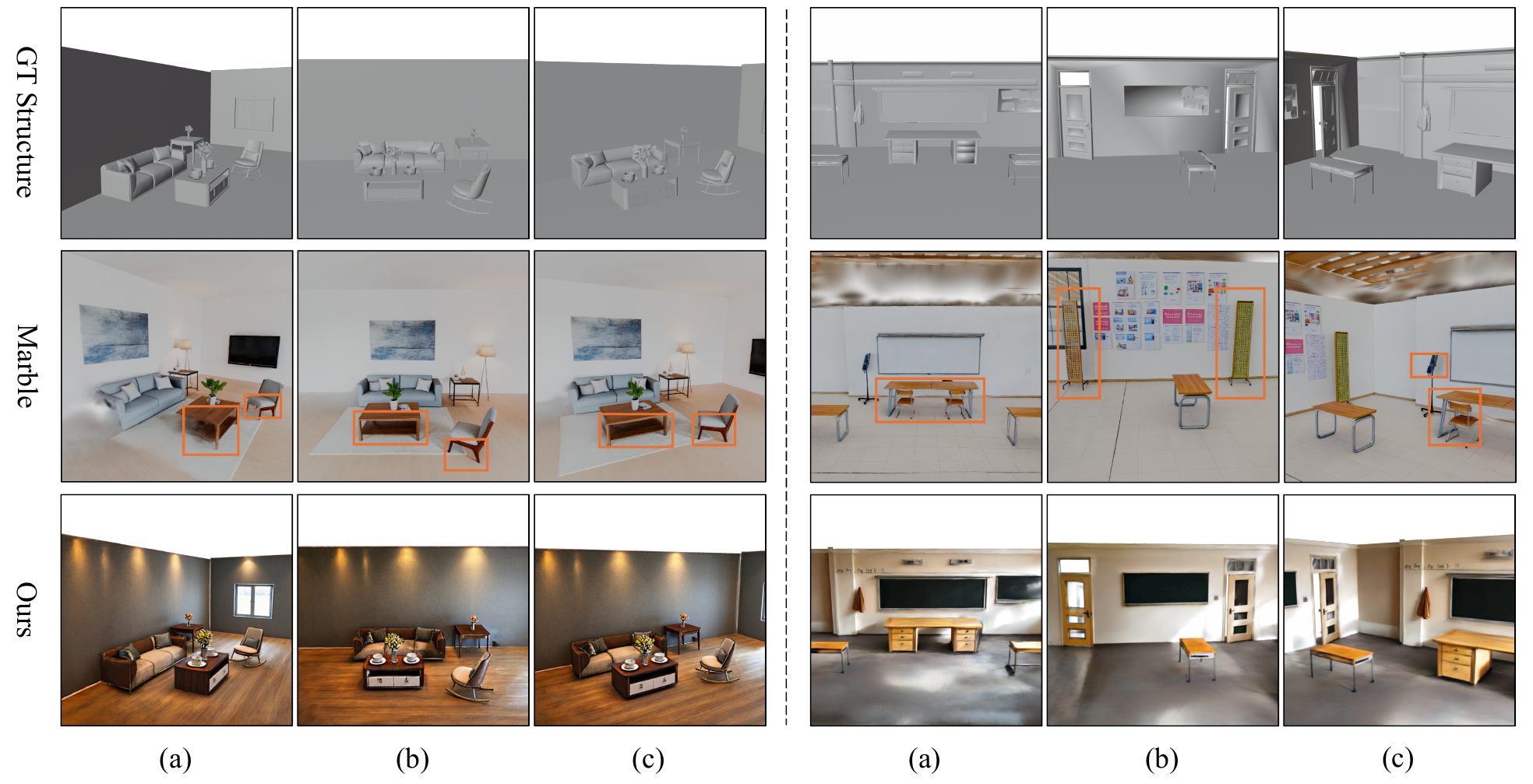}
  \caption{Qualitative comparisons with Marble~\cite{MarbleWorldLabs} (indoor pipeline). Orange boxes highlight Marble's geometric deviations.}
  \label{fig:comparison2}
\end{figure*}

\begin{figure*}
  \includegraphics[width=0.92 \textwidth, keepaspectratio]{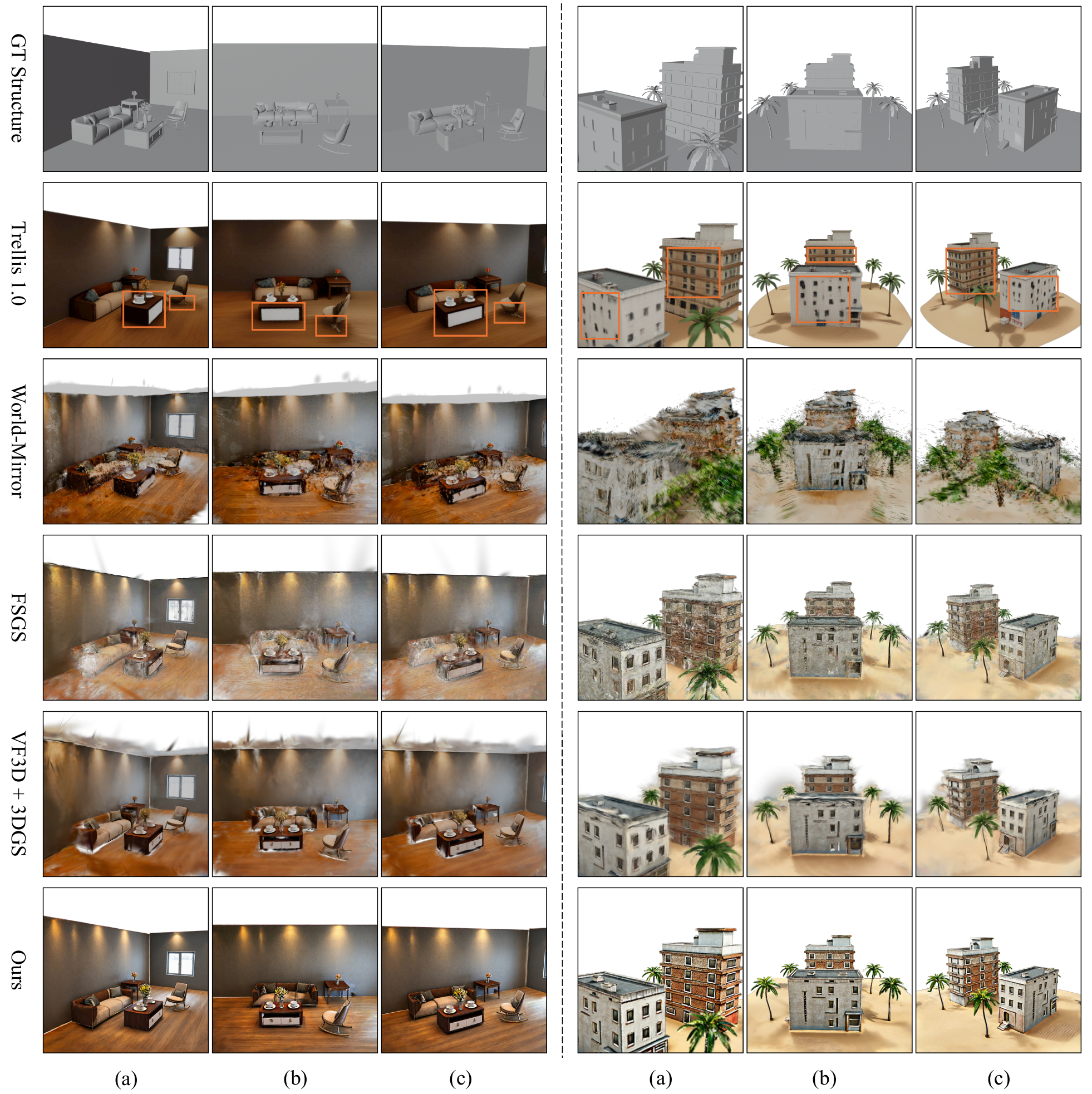}
  \caption{Qualitative comparisons with Trellis 1.0~\cite{xiang2024structured}, World-Mirror~\cite{huang2025voyager}, FSGS~\cite{zhu2023FSGS}, and VideoFrom3D~\cite{kim2025videofrom3d} (3DGS reconstructions from synthesized videos). Orange boxes highlight Trellis 1.0's geometric deviations.}
  \label{fig:comparison1}
\end{figure*}

~\cref{fig:comparison1} shows qualitative comparisons against the baseline methods.
Due to the limited geometric controllability of generative models, Trellis 1.0 often fails to preserve the original geometry when conditioned on existing 3D structures.
Traditional 3DGS reconstruction based on VideoFrom3D outputs and FSGS often suffer from needle-like artifacts and blurred details caused by view inconsistency.
World-Mirror relies on a VGGT~\cite{Wang_2025_CVPR}like prediction framework, where accurate geometric reconstruction becomes challenging when the inputs are inconsistent.
In ~\cref{fig:comparison2}, we show a separate comparison with Marble (indoor), which, similar to Trellis 1.0, often fails to preserve the input geometry.
In contrast, our method effectively avoids these issues and consistently achieves superior visual quality and higher structural fidelity, even under complex scene conditions.

\textit{Quantitative Comparisons.}
For the test dataset, we construct 12 3D scene models, including several scenes from Minecraft assets and manually created white-mesh scenes.
The white-mesh assets are collected from open asset platforms such as TurboSquid~\cite{turbosquid} and Free3D~\cite{free3d}. 
The dataset consists of three indoor scenes, four buildings-centric scenes and five landscape scenes. For each scene, we select two different text or style prompts, constructing a total of 24 test cases.
To evaluate our method against the baseline methods, we randomly sample 8 or 16 camera viewpoints per scene and report the following evaluation metrics.

For visual quality comparison, we use MUSIQ~\cite{Ke_2021_ICCV} and MANIQA~\cite{Yang_2022_CVPR}, which measure texture sharpness and perceptual realism from a human visual perspective.
For geometric consistency, we adopt PSNR-D, which measures the PSNR between ground-truth depth maps and depth maps estimated from rendered images using a monocular depth estimator.
For style consistency, following WonderWorld~\cite{Yu_2025_CVPR}, we compute the CLIP score between each rendered image and its corresponding pseudo-GT, and the cosine similarity between the CLIP embeddings of each novel view and the corresponding center view. ~\cref{tab:table1,tab:table2} shows quantitative comparisons with the baselines, indicating that our method consistently achieves better visual quality and stronger geometric consistency.

\begin{figure}[!t]
  \includegraphics[width=\columnwidth, keepaspectratio]{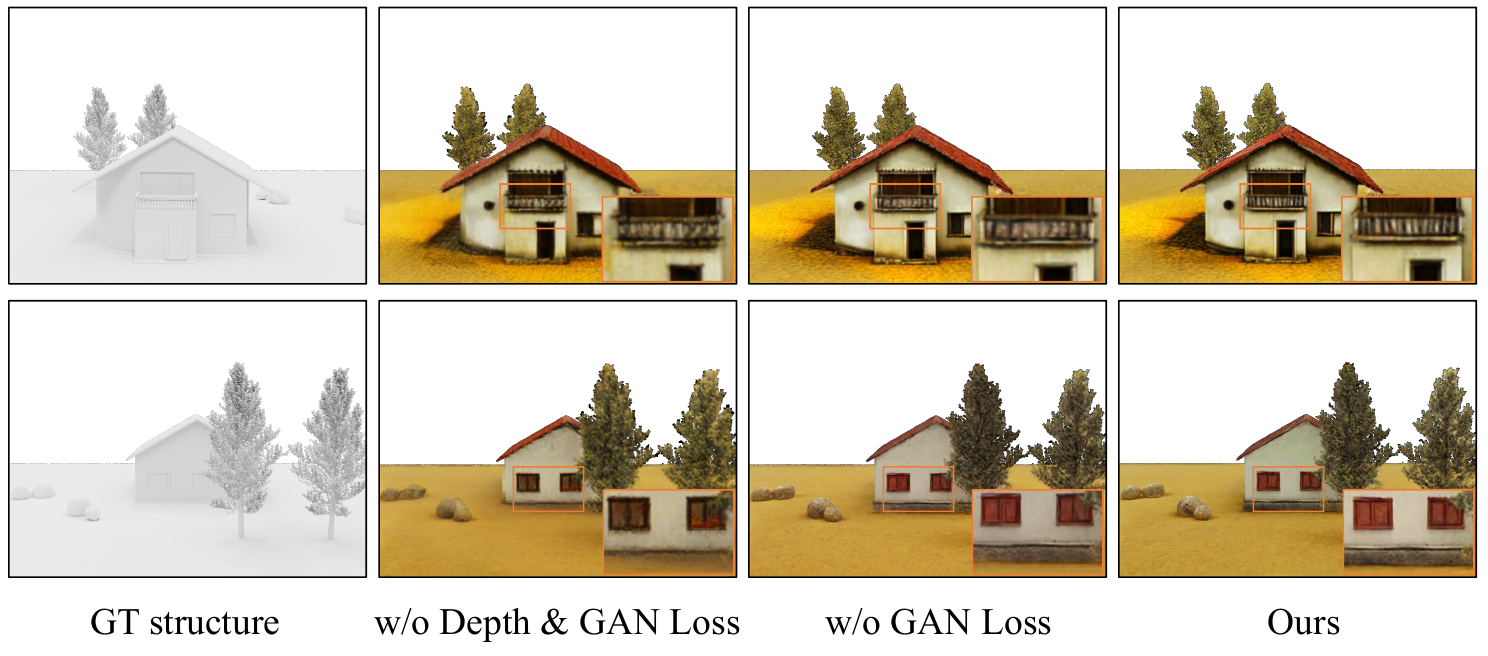}
  \vspace{-6mm}
  \caption{Qualitative results on the effectiveness of our self-supervised optimization strategy. Orange boxes indicate the same regions of the input geometry.}
  \label{fig:aba1}
\end{figure}

\begin{table}[!t]
\belowrulesep=0pt
\aboverulesep=0pt
\centering
\renewcommand\arraystretch{1.3}
\caption{Quantitative comparison of different optimization strategies.}
\vspace{-3mm}
\setlength{\tabcolsep}{3pt} 
\begin{tabular}{c|cccc}
\toprule
Method & PSNR-D $\uparrow$ & MUSIQ $\uparrow$ & MANIQ $\uparrow$ & CC $\uparrow$ \\
\midrule
w/o Depth \& GAN Loss & 12.67 & 34.74 & 0.29 & 0.88 \\
w/o GAN Loss & 14.67 & 40.31 & 0.30 & 0.89 \\
Ours & \textbf{19.00} & \textbf{57.08} & \textbf{0.33} & \textbf{0.93} \\
\bottomrule
\end{tabular}
\label{tab:table3}
\end{table}

\begin{figure}[!t]
  \includegraphics[width=0.8 \columnwidth, keepaspectratio]{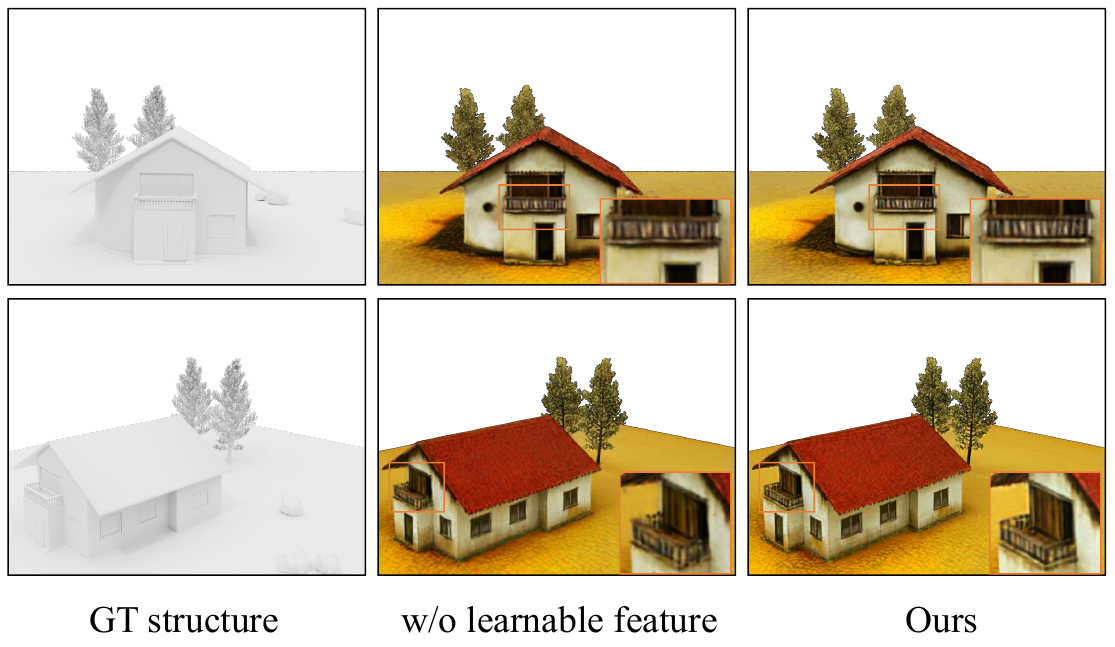}
  \vspace{-3mm}
  \caption{Qualitative results with and without the learnable features. Orange boxes indicate the same regions of the input geometry.}
  \label{fig:aba2}
\end{figure}

\begin{table}[!t]
\belowrulesep=0pt
\aboverulesep=0pt
\centering
\renewcommand\arraystretch{1.3}
\caption{Quantitative comparison of the impact of the learnable features.}
\vspace{-3mm}
\setlength{\tabcolsep}{3pt} 
\begin{tabular}{c|cccc}
\toprule
Method & PSNR-D $\uparrow$ & MUSIQ $\uparrow$ & MANIQ $\uparrow$ & CC $\uparrow$ \\
\midrule
w/o learnable feature & 17.82 & 45.55 & 0.31 & 0.92 \\
Ours & \textbf{19.00} & \textbf{57.08} & \textbf{0.33} & \textbf{0.93} \\
\bottomrule
\end{tabular}
\label{tab:table4}
\end{table}

\subsection{Ablation Study}

\indent \indent \textit{Effectiveness of Self-Supervised Optimization.}
We further investigate the effectiveness of our self-supervised optimization strategy through three ablation studies, as shown in ~\cref{fig:aba1}, which evaluate the impact of the depth loss, the GAN loss, and their joint supervision.
Removing both the depth and GAN losses and relying solely on the consistency loss (the second column of ~\cref{fig:aba1}) leads to noticeable structural artifacts, broken multi-view continuity, and blurred details, due to the lack of explicit structural constraints and detail refinement.
Removing only the GAN loss (the third column of ~\cref{fig:aba1}) significantly improves the structural coherence of the scene but results in insufficient visual richness and texture fidelity.
Using the full self-supervised optimization strategy (the fourth column of ~\cref{fig:aba1}) effectively balances structural regularity and visual richness.
The quantitative results reported in ~\cref{tab:table3} further demonstrate the effectiveness of the strategies we proposed.  

\indent \indent \textit{Effectiveness of Learnable Features.}
The initial voxel features are obtained by back-projecting and then averaging multi-view features.
While this process alleviates cross-view inconsistencies, it also weakens the representation of local high-frequency information.
The qualitative results in ~\cref{fig:aba2} demonstrate the effectiveness of the proposed learnable features: the reconstructed balcony railings are noticeably clearer across multiple viewpoints, exhibiting richer geometric and texture details.
This improvement arises because the learnable features are continuously updated during training, enabling the model to adaptively capture and recover missing local details. 
The quantitative results reported in ~\cref{tab:table4} further support this observation, showing consistent improvements across multiple reconstruction quality metrics. 
Together, these results confirm the crucial role of learnable features in enhancing detail recovery and overall reconstruction quality, achieving a smooth transition from coarse-grained to fine-grained 3D reconstruction. See the supplementary material for more ablation experiments.

\section{Conclusion}
\label{sec6:conclusion}
We propose GeoDiff3D, a geometry-guided generative framework that leverages coarse geometry as a structural anchor and diffusion models as a strong prior for detail synthesis to efficiently generate high-quality 3D scenes.
Through three stages—texture reference image generation, voxel-aligned 3D feature aggregation, and self-supervised optimization—our framework effectively exploits the structural anchoring of coarse geometry and the rich detail generation capability of 2D diffusion models.
The voxel-aligned mechanism and self-supervised strategy effectively balance structural coherence and visual fidelity, while significantly reducing reliance on large-scale labeled data and extensive computational resources.
Extensive experiments demonstrate that GeoDiff3D achieves robust performance on complex scenes, outperforming existing baselines in both generation quality and adaptability, providing a low-barrier and efficient solution for practical 3D scene construction.

\indent \indent \textit{Limitations.}
Our method does not explicitly model geometry-free regions, which limits the realism of sky and atmospheric effects. Moreover, using line-drawing cues as diffusion priors cannot capture continuous depth changes, weak textures, or complex occlusions, making pseudo-GT guidance unreliable in challenging cases. Future work will explore explicit sky/background modeling and richer geometric cues (e.g., depth/normal/semantic signals) to improve robustness. More detailed information can be found in the supplementary material.

\bibliographystyle{ACM-Reference-Format}
\bibliography{sample-bibliography}

\end{document}